\documentclass{amsart}
\usepackage{amssymb}
\usepackage{amsfonts}

\newtheorem{theorem}{Theorem}
\theoremstyle{plain}

\newtheorem{definition}{Definition}
\newtheorem{example}{Example}

\newtheorem{lemma}{Lemma}

\newtheorem{proposition}{Proposition}
\newtheorem{remark}{Remark}

\numberwithin{equation}{section}
\input{tcilatex}

\begin{document}
\title[Bipolar Fuzzy Soft sets and Its Applications]{Bipolar Fuzzy Soft sets
and its applications in decision making problem}
\author{Muhammad Aslam}
\address{Department of Mathematics, Quaid-i-Azam University 45320 Islamabad
44000, Pakistan}
\email{draslamqau@yahoo.com}
\author{Saleem Abdullah}
\address{Department of Mathematics, Quaid-i-Azam University 45320 Islamabad
44000, Pakistan}
\email{saleemabdullah81@yahoo.com}
\author{Kifayat ullah}
\address{Department of Mathematics, Quaid-i-Azam University 45320 Islamabad
44000, Pakistan}

\begin{abstract}
In this article, we combine the concept of a bipolar fuzzy set and a soft
set. We introduce the notion of bipolar fuzzy soft set and study fundamental
properties. We study basic operations on bipolar fuzzy soft set. We define
exdended union, intersection of two bipolar fuzzy soft set. We also give an
application of bipolar fuzzy soft set into decision making problem. We give
a general algorithm to solve decision making problems by using bipolar fuzzy
soft set.
\end{abstract}

\maketitle

\section{Introduction}

Complicated problems in different field like engineering, economics,
environmental science, medicine and social sciences, which arising due to
classical mathematical modelling and manipolating of varouis type of
uncertainty. While some of traditional mathematical tool fail to solve these
complicated problems. We used some mathematical modelling like fuzzy set
theory \cite{a}, rough set theory \cite{b}, interval mathematics [12] and
probability theory are well-known and operative tools for handling with
vagueness and uncertainty, each of them has its own inherent limitations;
one major fault shared by these mathematical methodologies may be due to the
inadequacy of parametrization tools \cite{d}.

Molodtsov, \cite{d} adopted the notion of soft sets. Soft set is a new
mathematical tool to describe the uncertainties. Soft set theory is powerful
tool to describe uncertainties. Recently, researcher are engaged in soft set
theory. Maji et al. \cite{f} defined new notions on soft sets. Ali et al. 
\cite{ali} studied some new concepts of a soft set. Sezgin and Atag\"{u}n 
\cite{op} studied some new theoretical soft set operations. Majumdar and
Samanta, worked on soft mappings \cite{majum}. Choudhure et al. defined the
concept of soft relation and fuzzy soft relation and then applied them to
solve a number of decision- making problems. In \cite{2}, Akta\d{s} and \d{C}%
a\u{g}man applied the concept of soft set to groups theory and adopted soft
group of a group. Feng et.al, studied and applied softness to semirings\cite%
{3}. Recently, Acar studied soft rings \cite{4}. Jun et. al, applied the
concept of soft set to BCK/BCI-algebras \cite{jun1,jun2,jun3}. Sezgin and
Atag\"{u}n initiated th concept of normalistic soft groups \cite{op1}. Zhan
et.al, worked on soft ideal of BL-algebras \cite{zhan}. In \cite{kazanc},
Kazanc\i\ et. al, used the concept of soft set to BCH-algebras. Sezgin et.
al, studied soft nearrings \cite{sez-near2}. \d{C}a\u{g}man et al.
considered two types of notions of a soft set with group, which is called
group Soft intersection group softunion groups of a group \cite{cag-fil}.
see \cite{AWE}.

Fuzzy set originally proposed by Zadeh in \cite{a} of 1965. After semblance
of the concept of fuzzy set, researcher given much atttention to developed
fuzzy set theory. Maji et al. \cite{7} introduced the concept of fuzzy soft
sets. Afterwards, many researchers have worked on this concept. Roy and Maji 
\cite{ASE} provided some results on an application of fuzzy soft sets in
decision making problems. F. Feng et al. give application in decision making
problem \cite{f1,f2}

Fuzzy set is a type of important mathematical structure to represent a
collection of objects whose boundary is vague. There are several types of
fuzzy set extensions in the fuzzy set theory, for example, intuitionistic
fuzzy sets, interval-valued fuzzy sets, vague sets, etc. bi-polar-valued
fuzzy set is another an extension of fuzzy set whose membership degree range
is different from the above extensions. In $2000$, Lee \cite{5} initiated an
extension of fuzzy set named bi-polar-valued fuzzy set. He gave two kinds of
representations of the notion of ni-polar-valued fuzzy sets. In case of
Bi-polar-valued fuzzy sets membership degree range is enlarged from the
interval $[0,1]$ to $[-1,1]$. In a bi-polar-valued fuzzy set, the membership
degree 0 indicate that elements are irrelevant to the corresponding
property, the membership degrees on $(0,1]$ assigne that elements some what
satisfy the property, and the membership degrees on $[-1,0)$ assigne that
elements somewhat satisfy the implicit counter-property \cite{5}.

In this article, we combine the concept of a bipolar fuzzy set and a soft
set. We introduce the notion of bipolar fuzzy soft set and study fundamental
properties. We study basic operations on bipolar fuzzy soft set. We define
exdended union, intersection of two bipolar fuzzy soft set. We also give an
application of bipolar fuzzy soft set into decision making problem. We give
a general algorithm to solve decision making problems by using bipolar fuzzy
soft set.

\section{Preliminaries}

In this section we provide previous concept of bipolar fuzzy sets, soft sets
and fuzzy soft sets.

\begin{definition}
\cite{5}A bipolar fuzzy set $A$ in a universe $U$ is an object having the
form, $A=\{(x,%
{\mu}%
_{A}^{+}(x),%
{\mu}%
_{A}^{-}(x)):x\in U\}$ where $%
{\mu}%
_{A}^{+}:U\rightarrow \lbrack 0,1]$, $%
{\mu}%
_{A}^{-}:U\rightarrow \lbrack -1,0]$. So $%
{\mu}%
_{A}^{+}$ denote for positive information and $%
{\mu}%
_{A}^{-}$ denote for negative information.
\end{definition}

\begin{definition}
\cite{d}Let $U$ be an initial universe, $E$ be the set of parameters, $%
A\subset E$ and $P(U)$ is the power set of $U$. Then $(F,A)$ is called a soft%
\textbf{\ }set, where $F:A\rightarrow P(U)$.
\end{definition}

\begin{definition}
\cite{ali}For two soft sets $(F,A)$ and $(G,B)$ over a common universe $U$%
,we say that $(F,A)$ is a soft subset of $(G,B)$, denoted by $(F,A)\tilde{%
\subset}(G,B)$, if it satisfies.
\end{definition}

\begin{enumerate}
\item $A\subset B$

\item $\forall $ $a\in A$, $F(a)$ is a subset of $G(a)$.
\end{enumerate}

Similarly, $(F,A)$ is called a superset of $(G,B)$ if $(G,B)$ is a soft
subset of $(F,A)$. This relation is denoted by $(F,A)\tilde{\supset}(G,B)$.

\begin{definition}
\cite{f} If $(F,A)$ and $(G,B)$ are two soft sets over a common universe U.
The union of $(F,A)$ and $(G,B)$ is defined to be the soft set $(H,C)$
satisfying the following conditions: (i) $C=A\cup B$: (ii) for all $c\in C,$%
\begin{eqnarray*}
H(c) &=&F(c)\text{ if }c\in A\backslash B \\
&=&G(c)\text{ if }c\in B\backslash A \\
&=&F(c)\cup G(c)\text{ if }c\in A\cap B
\end{eqnarray*}%
This relation is denoted by $(H,C)=(F,A)\tilde{\cup}(G,B)$.
\end{definition}

\begin{definition}
\cite{ali}Let $(F,A)$ and $(G,B)$ be two soft sets over a common universe $U$
such that $A\cap B\neq \emptyset $. The restricted intersection of $(F,A)$
and $(G,B)$ is defined to be the soft set $(H,C),$ $C=A\cap B$ and $\forall $
$c\in C$, $H(c)=F(c)\cap G(c)$. We write $(H,C)=(F,A)\Cap (G,B)$.
\end{definition}

\begin{definition}
\cite{7} Let $U$ be an initial universe, $E$ be the set of all parameters, $%
A\subset E$ and $\tilde{P}(U)$ is the collection of all fuzzy subsets of $U$%
. Then $(F,A)$ is called fuzzy soft\textbf{\ }set, where $F:A\rightarrow 
\tilde{P}(U)$.
\end{definition}

\begin{definition}
\cite{7} If $(F,A)$ and $(G,B)$ are two fuzzy soft sets over a common
universe $U$, then the union of $(F,A)$ and $(G,B)$ is defined to be the
fuzzy soft set $(H,C)$ satisfying the following conditions: (i) $C=A\cup B$:
(ii) $c\in C$,%
\begin{eqnarray*}
H(c) &=&F(c)\text{ if }c\in A\backslash B\text{ \ } \\
&=&G(c)\text{ if }c\in B\backslash A \\
&=&F(c)\cup G(c)\text{ if }c\in A\cap B
\end{eqnarray*}%
This relation is denoted by$(H,C)=(F,A)\tilde{\cup}(G,B)$.
\end{definition}

\section{Bipolar Fuzzy Soft Sets.}

In this section we introduce the concept of bipolar fuzzy soft set, absolute
bipolar fuzzy soft set, null bipolar fuzzy soft set and complement of
bipolar fuzzy soft set

\begin{definition}
Let $U$ be a universe, $E$ a set of parameters and $A\subset E$. Define $%
F:A\rightarrow BF^{U}$, where $BF^{U}$ is the collection of all bipolar
fuzzy subsets of $U$. Then $\left( F,A\right) $ is said to be a bipolar
fuzzy soft set over a universe $U$. It is defined by%
\begin{equation*}
(F,A)=\{(x,\text{ }\mu _{e}^{+}(x),\text{ }\mu _{e}^{-}(x)\text{ })\text{:
for all }x\in U\text{ and }e\in A\}
\end{equation*}
\end{definition}

\begin{example}
Let $U=$\{$c_{1}$, $c_{2}$, $c_{3}$, $c_{4}$\} be the set of four cars under
consideration and $E=$\{$e_{1}$=Costly, $e_{2}$=Beautiful, $e_{3}$=Fuel
Efficient, $e_{4}$=Modern Technology \} be the set of parameters and $A=$\{$%
e_{1},e_{2},e_{3}$\}$\subseteq E$. Then,%
\begin{equation*}
\left( F,A\right) =\left\{ 
\begin{array}{c}
F\left( e_{1}\right) =\left\{ 
\begin{array}{c}
\left( c_{1},0.1,-0.5\right) ,\left( c_{2},0.3,-0.6\right) , \\ 
\left( c_{3},0.4,-0.2\right) ,\left( c_{4},0.7,-0.2\right)%
\end{array}%
\right\} , \\ 
F\left( e_{2}\right) =\left\{ 
\begin{array}{c}
\left( c_{1},0.3,-0.5\right) ,\left( c_{2},0.4,-0.2\right) , \\ 
\left( c_{3},0.5,-0.2\right) ,\left( c_{4},0.4,-0.2\right)%
\end{array}%
\right\} , \\ 
F\left( e_{3}\right) =\left\{ 
\begin{array}{c}
\left( c_{1},0.8,-0.11\right) ,\left( c_{2},0.3,-0.6\right) , \\ 
\left( c_{3},0.4,-0.3\right) ,\left( c_{4},0.6,-0.2\right)%
\end{array}%
\right\}%
\end{array}%
\right\}
\end{equation*}
\end{example}

\begin{definition}
Let $U$ be a universe and $E$ a set of attributes.Then, $\left( U,E\right) $
is the collection of all bipolar fuzzy soft sets on $U$ with attributes from 
$E$ and is said to be bipolar fuzzy soft class.
\end{definition}

\begin{definition}
A bipolar fuzzy soft set $\left( F,A\right) $ is said to be a null bipolar
fuzzy soft set denoted by empty set $\emptyset $, if for all $e\in A$, $%
F\left( e\right) =\emptyset .$
\end{definition}

\begin{definition}
A bipolar fuzzy soft set $\left( F,A\right) $ is said to be an absolute
bipolar fuzzy soft set. If for all $e\in A$, $F\left( e\right) =BF^{U}$
\end{definition}

\begin{definition}
The complement of a bipolar fuzzy soft set $\left( F\text{, }A\right) $ is
denoted $\left( F\text{, }A\right) ^{c}$ and is defined by $\left( F\text{, }%
A\right) ^{c}=\{(x,1-%
{\mu}%
_{A}^{+}\left( x\right) ,$ $-1-%
{\mu}%
_{A}^{-}\left( x\right) ):x\in U\}$.
\end{definition}

\begin{example}
Let $U=$\{$b_{1}$, $b_{2}$, $b_{3}$, $b_{4}$\} be the set of four bikes
under consideration and $E=$\{ $e_{1}=$Stylish, $e_{2}=$Heavy duty, $e_{3}=$%
Light, $e_{4}=$Steel \} be the set of parameters and $A=$\{$e_{1}$, $e_{2}$%
\} be subset of $E$. Then,%
\begin{equation*}
\left( F,A\right) =\left\{ 
\begin{array}{c}
F\left( e_{1}\right) =\left\{ 
\begin{array}{c}
\left( b_{1},0.1,-0.5\right) ,\left( b_{2},0.3,-0.6\right) \\ 
\left( b_{3},0.4,-0.2\right) ,\left( b_{4},0.7,-0.2\right)%
\end{array}%
\right\} , \\ 
F\left( e_{2}\right) =\left\{ 
\begin{array}{c}
\left( b_{1},0.3,-0.5\right) ,\left( b_{2},0.4,-0.2\right) , \\ 
\left( b_{3},0.5,-0.2\right) ,\left( b_{4},0.4,-0.2\right)%
\end{array}%
\right\}%
\end{array}%
\right\}
\end{equation*}%
The complement of the bipolar fuzzy soft set $\left( F\text{,}A\right) $ is 
\begin{equation*}
\left( F,A\right) ^{c}=\left\{ 
\begin{array}{c}
F\left( e_{1}\right) =\left\{ 
\begin{array}{c}
\left( b_{1},0.9,-0.5\right) ,\left( b_{2},0.7,-0.4\right) \\ 
\left( b_{3},0.6,-0.8\right) ,\left( b_{4},0.3,-0.8\right)%
\end{array}%
\right\} , \\ 
F\left( e_{2}\right) =\left\{ 
\begin{array}{c}
\left( b_{1},0.7,-0.5\right) ,\left( b_{2},0.6,-0.8\right) , \\ 
\left( b_{3},0.5,-0.8\right) ,\left( b_{4},0.6,-0.8\right)%
\end{array}%
\right\}%
\end{array}%
\right\}
\end{equation*}
\end{example}

\section{ Bipolar Fuzzy Soft Subsets}

\begin{definition}
Let $\left( F,A\right) $ and $\left( G,B\right) $ be two bipolar fuzzy soft
sets over a common universe $U$. We say that $\left( F,A\right) $ is a
bipolar fuzzy soft subset of $\left( G,B\right) $, if $\left( 1\right) $ $%
A\subseteq B$ and $\left( 2\right) $ \ $\forall $ $e\in A$, $F\left(
e\right) $ is a bipolar fuzzy subset of $G\left( e\right) $. We write $%
\left( F,A\right) \bar{\subset}\left( G,B\right) $.
\end{definition}

\begin{remark}
Every element of $\left( F,A\right) $ is presented in $\left( G,B\right) $
and do not depend on its membership or non-membership.
\end{remark}

\begin{example}
Let $U=$\{$m_{1}$, $m_{2}$, $m_{3}$, $m_{4}$\} be the set of four men under
consideration and $E=$\{ $e_{1}=$ Educated, $e_{2}=$ Government employee, $%
e_{3}=$ Businessman, $e_{4}=$ Smart \} be the set of parameters and $A=$\{$%
e_{1}$, $e_{2}$ \}, $B=$\{ $e_{1}$, $e_{2}$, $e_{3}$ \} be subsets of $E$.
Then,%
\begin{equation*}
\left( F,A\right) =\left\{ 
\begin{array}{c}
F\left( e_{1}\right) =\left\{ 
\begin{array}{c}
\left( m_{1},0.1,-0.5\right) ,\left( m_{2},0.3,-0.6\right) \\ 
\left( m_{3},0.4,-0.2\right) ,\left( m_{4},0.7,-0.2\right)%
\end{array}%
\right\} , \\ 
F\left( e_{2}\right) =\left\{ 
\begin{array}{c}
\left( m_{1},0.3,-0.5\right) ,\left( m_{2},0.4,-0.2\right) , \\ 
\left( m_{3},0.5,-0.2\right) ,\left( m_{4},0.4,-0.2\right)%
\end{array}%
\right\}%
\end{array}%
\right\}
\end{equation*}%
\begin{equation*}
\left( G,B\right) =\left\{ 
\begin{array}{c}
G\left( e_{1}\right) =\left\{ 
\begin{array}{c}
\left( m_{1},0.2,-0.5\right) ,\left( m_{2},0.2,-0.6\right) , \\ 
\left( m_{3},0.2,-0.3\right) ,\left( m_{4},0.7,-0.1\right)%
\end{array}%
\right\} , \\ 
G\left( e_{2}\right) =\left\{ 
\begin{array}{c}
\left( m_{1},0.3,-0.6\right) ,\left( m_{2},0.2,-0.5\right) , \\ 
\left( m_{3},0.5,-0.3\right) ,\left( m_{4},0.5,-0.2\right)%
\end{array}%
\right\} , \\ 
G\left( e_{3}\right) =\left\{ 
\begin{array}{c}
\left( m_{1},0.8,-0.01\right) ,\left( m_{2},0.4,-0.6\right) , \\ 
\left( m_{3},0.2,-0.3\right) ,\left( m_{4},0.7,-0.2\right)%
\end{array}%
\right\}%
\end{array}%
\right\}
\end{equation*}%
$A\subseteq B$ and for all $e\in A$, $F\left( e\right) \leq G\left( e\right) 
$. Then $\left( F,A\right) \bar{\subset}\left( G,B\right) .$
\end{example}

\begin{definition}
Let $\left( F,A\right) $ and $\left( G,B\right) $ be two bipolar fuzzy soft
sets over a common universe $U$. We say that $\left( F,A\right) $ and $%
\left( G,B\right) $ are bipolar fuzzy soft \ equal sets if $\left(
F,A\right) $ is a bipolar fuzzy soft subset of $\left( G,B\right) $ and $%
\left( G,B\right) $ is a bipolar fuzzy soft subset of $\left( F,A\right) $.
\end{definition}

\section{Operations\ on Bipolar Fuzzy Soft Sets}

\begin{definition}
An intersection of two bipolar fuzzy soft sets $\left( F,A\right) $ and $%
\left( G,B\right) $ is a bipolar fuzzy soft set $\left( H,C\right) $, where $%
C=A\cap B\neq \emptyset $ and $H:C\rightarrow BF^{U}$ is defined by $H\left(
e\right) =F\left( e\right) \cap G\left( e\right) $ $\forall $ $e\in C$ and \
denoted by $\left( H,C\right) =\left( F,A\right) \bar{\cap}\left( G,B\right) 
$.
\end{definition}

\begin{example}
Let $U=$\{$b_{1}$, $b_{2}$, $b_{3}$, $b_{4}$\} be the set of four bikes
under consideration and $E=$\{$e_{1}$=Light, $e_{2}$=Beautiful, $e_{3}$=Good
millage, $e_{4}$=Modern Technology \} be the set of parameters and $A=$\{$%
e_{1}$, $e_{2}$ \}$\subseteq E$, $B=$\{$e_{1}$, $e_{2}$, $e_{3}$\}$\subseteq
E$. Then,%
\begin{equation*}
\left( F,A\right) =\left\{ 
\begin{array}{c}
F\left( e_{1}\right) =\left\{ 
\begin{array}{c}
\left( b_{1},0.1,-0.5\right) ,\left( b_{2},0.3,-0.6\right) , \\ 
\left( b_{3},0.4,-0.2\right) ,\left( b_{4},0.7,-0.2\right)%
\end{array}%
\right\} , \\ 
F\left( e_{2}\right) =\left\{ 
\begin{array}{c}
\left( b_{1},0.3,-0.5\right) ,\left( b_{2},0.4,-0.2\right) , \\ 
\left( b_{3},0.4,-0.4\right) ,\left( b_{4},0.4,-0.2\right)%
\end{array}%
\right\}%
\end{array}%
\right\}
\end{equation*}%
\begin{equation*}
\left( G,B\right) =\left\{ 
\begin{array}{c}
G\left( e_{1}\right) =\left\{ 
\begin{array}{c}
\left( b_{1},0.2,-0.5\right) ,\left( b_{2},0.2,-0.6\right) , \\ 
\left( b_{3},0.2,-0.3\right) ,\left( b_{4},0.7,-0.1\right)%
\end{array}%
\right\} , \\ 
G\left( e_{2}\right) =\left\{ 
\begin{array}{c}
\left( b_{1},0.3,-0.6\right) ,\left( b_{2},0.2,-0.5\right) , \\ 
\left( b_{3},0.5,-0.3\right) ,\left( b_{4},0.5,-0.2\right)%
\end{array}%
\right\} , \\ 
G\left( e_{3}\right) =\left\{ 
\begin{array}{c}
\left( b_{1},0.8,-0.01\right) ,\left( b_{2},0.4,-0.6\right) , \\ 
\left( b_{3},0.2,-0.3\right) ,\left( b_{4},0.7,-0.2\right)%
\end{array}%
\right\}%
\end{array}%
\right\}
\end{equation*}%
Then $\left( H,C\right) =\left( F,A\right) \bar{\cap}\left( G,B\right) $,
where $C=A\cap B=$\{$e_{1}$, $e_{2}$\}%
\begin{equation*}
\left( H,C\right) =\left\{ 
\begin{array}{c}
H\left( e_{1}\right) =\left\{ 
\begin{array}{c}
\left( b_{1},0.1,-0.5\right) ,\left( b_{2},0.2,-0.6\right) , \\ 
\left( b_{3},0.2,-0.2\right) ,\left( b_{4},0.7,-0.1\right)%
\end{array}%
\right\} , \\ 
H\left( e_{2}\right) =\left\{ 
\begin{array}{c}
\left( b_{1},0.3,-0.5\right) ,\left( b_{2},0.2,-0.2\right) , \\ 
\left( b_{3},0.4,-0.3\right) ,\left( b_{4},0.4,-0.2\right)%
\end{array}%
\right\}%
\end{array}%
\right\}
\end{equation*}
\end{example}

\begin{definition}
Union of two bipolar fuzzy soft sets over a common universe $U$ is a bipolar
fuzzy soft set $\left( H,C\right) $, where $C=A\cup B$ and $H:C\rightarrow
BF^{U\text{ }}$ is defined by%
\begin{eqnarray*}
H\left( e\right) &=&F\left( e\right) \text{ if }e\in A\setminus B \\
&=&G\left( e\right) \text{ if }e\in B\setminus A \\
&=&F\left( e\right) \cup G\left( e\right) \text{ if }e\in A\cap B
\end{eqnarray*}%
and \ denoted by $\left( H,C\right) =\left( F,A\right) \bar{\cup}\left(
G,B\right) $.
\end{definition}

\begin{example}
Let $U=$\{$c_{1}$, $c_{2}$, $c_{3}$, $c_{4}$\} be the set of four cars under
consideration and $E=$\{$e_{1}=$Costly, $e_{2}=$Beautiful, $e_{3}=$Fuel
Efficient, $e_{4}=$Modern Technology \} be the set of parameters and $A=$\{$%
e_{1}$, $e_{2}$, $e_{3}$\}$\subseteq E$, $B=$\{$e_{1}$, $e_{2}$, $e_{3}$, $%
e_{4}$\}$\subseteq E$. Then%
\begin{equation*}
\left( F,A\right) =\left\{ 
\begin{array}{c}
F\left( e_{1}\right) =\left\{ 
\begin{array}{c}
\left( c_{1},0.1,-0.5\right) ,\left( c_{2},0.3,-0.6\right) , \\ 
\left( c_{3},0.4,-0.2\right) ,\left( c_{4},0.7,-0.2\right)%
\end{array}%
\right\} , \\ 
F\left( e_{2}\right) =\left\{ 
\begin{array}{c}
\left( c_{1},0.3,-0.5\right) ,\left( c_{2},0.4,-0.2\right) , \\ 
\left( c_{3},0.5,-0.2\right) ,\left( c_{4},0.4,-0.2\right)%
\end{array}%
\right\} , \\ 
F\left( e_{3}\right) =\left\{ 
\begin{array}{c}
\left( c_{1},0.8,-0.1\right) ,\left( c_{2},0.3,-0.6\right) , \\ 
\left( c_{3},0.4,-0.3\right) ,\left( c_{4},0.6,-0.2\right)%
\end{array}%
\right\}%
\end{array}%
\right\}
\end{equation*}%
\begin{equation*}
\left( G,B\right) =\left\{ 
\begin{array}{c}
G\left( e_{1}\right) =\left\{ 
\begin{array}{c}
\left( c_{1},0.2,-0.5\right) ,\left( c_{2},0.2,-0.6\right) , \\ 
\left( c_{3},0.2,-0.3\right) ,\left( c_{4},0.7,-0.1\right)%
\end{array}%
\right\} , \\ 
G\left( e_{2}\right) =\left\{ 
\begin{array}{c}
\left( c_{1},0.3,-0.6\right) ,\left( c_{2},0.2,-0.5\right) , \\ 
\left( c_{3},0.5,-0.3\right) ,\left( c_{4},0.5,-0.2\right)%
\end{array}%
\right\} , \\ 
G\left( e_{3}\right) =\left\{ 
\begin{array}{c}
\left( c_{1},0.8,-0.01\right) ,\left( c_{2},0.4,-0.6\right) \\ 
,\left( c_{3},0.2,-0.3\right) ,\left( c_{4},0.7,-0.2\right)%
\end{array}%
\right\} \\ 
G\left( e_{4}\right) =\left\{ 
\begin{array}{c}
\left( c_{1},0.1,-0.6\right) ,\left( c_{2},0.3,-0.4\right) , \\ 
\left( c_{3},0.1,-0.6\right) ,\left( c_{4},0.0,-0.2\right)%
\end{array}%
\right\}%
\end{array}%
\right\}
\end{equation*}%
Then $\left( H,C\right) =\left( F,A\right) \bar{\cup}\left( G,B\right) $
,where $C=A\cup B=$\{$e_{1}$, $e_{2}$, $e_{3}$, $e_{4}$\}%
\begin{equation*}
\left( H,C\right) =\left\{ 
\begin{array}{c}
H\left( e_{1}\right) =\left\{ 
\begin{array}{c}
\left( c_{1},0.2,-0.5\right) ,\left( c_{2},0.3,-0.6\right) , \\ 
\left( c_{3},0.4,-0.3\right) ,\left( c_{4},0.7,-0.2\right)%
\end{array}%
\right\} , \\ 
H\left( e_{2}\right) =\left\{ 
\begin{array}{c}
\left( c_{1},0.3,-0.6\right) ,\left( c_{2},0.4,-0.5\right) , \\ 
\left( c_{3},0.5,-0.3\right) ,\left( c_{4},0.5,-0.2\right)%
\end{array}%
\right\} , \\ 
H\left( e_{3}\right) =\left\{ 
\begin{array}{c}
\left( c_{1},0.8,-0.1\right) ,\left( c_{2},0.4,-0.6\right) , \\ 
\left( c_{3},0.4,-0.3\right) ,\left( c_{4},0.7,-0.2\right)%
\end{array}%
\right\} \\ 
H\left( e_{4}\right) =\left\{ 
\begin{array}{c}
\left( c_{1},0.1,-0.6\right) ,\left( c_{2},0.3,-0.4\right) , \\ 
\left( c_{3},0.1,-0.6\right) ,\left( c_{4},0.0,-0.2\right)%
\end{array}%
\right\}%
\end{array}%
\right\}
\end{equation*}
\end{example}

\begin{definition}
Let $T=$\{$\left( F_{i},A_{i}\right) $ : $i\in I$\} be a family of bipolar
fuzzy soft sets in a bipolar fuzzy soft class $(U,E)$. Then the intersection
of bipolar fuzzy soft sets in $T$ is a bipolar fuzzy soft set $\left(
H,C\right) $, where $C=\cap A_{i}\neq \emptyset $ for all $i\in I$, $H\left(
e\right) =\cap F_{i}\left( e\right) $ \ for all $e\in C.$
\end{definition}

\begin{definition}
Let $T=$\{$\left( F_{i},A_{i}\right) $ : $i\in I$\} be a family of bipolar
fuzzy soft sets in a bipolar fuzzy soft class $(U,E)$. Then the union of
bipolar fuzzy soft sets in $T$ is a bipolar fuzzy soft set$\left( H,C\right) 
$, where $C=\cup A_{i}$ for all $i\in I.$%
\begin{eqnarray*}
H\left( e\right)  &=&F_{i}\left( e\right) \text{ if }e\in A_{i} \\
&=&\emptyset \text{ if }e\notin A_{i}
\end{eqnarray*}
\end{definition}

\begin{definition}
Let $\left( F\text{, }A\right) $\ and $\left( G\text{, }B\right) $ be two
bipolar fuzzy soft sets over a common universe $U$. The extended
intersection of $\left( F\text{,}A\right) $ and $\left( G\text{, }B\right) $
is defined t o be the\ bipolar fuzzy soft set $\left( H\text{,}C\right) $,
where $C=A\cap B$ and for all $e\in C.$%
\begin{eqnarray*}
H\left( e\right)  &=&F\left( e\right) \text{ if }e\in A\backslash B \\
&=&G\left( e\right) \text{ if }e\in B\backslash A \\
&=&F\left( e\right) \cap G\left( e\right) \text{ if }e\in A\cap B
\end{eqnarray*}%
This intersection is denoted by $\left( H\text{,}C\right) =\left( F\text{,}%
A\right) \sqcap _{\epsilon }\left( G\text{,}B\right) $.
\end{definition}

\begin{definition}
Let $\left( F\text{, }A\right) $\ and $\left( G\text{,}B\right) $ be two
bipolar fuzzy soft sets over a common universe $U$. The restricted union of $%
\left( F\text{,}A\right) $ and $\left( G\text{,}B\right) $ is defined to be
the\ bipolar fuzzy soft set $\left( H\text{,}C\right) $, where $C=A\cap
B\neq $ $\emptyset $ and for all $e\in C$%
\begin{equation*}
H\left( e\right) =F\left( e\right) \cup G\left( e\right)
\end{equation*}%
This union is denoted by $\left( H\text{,}C\right) =\left( F\text{,}A\right) 
\bar{\cup}_{R}\left( G\text{,}B\right) $.
\end{definition}

\begin{proposition}
Let $\left( F,A\right) $ be bipolar fuzzy soft set over a common universe $U$%
. Then,
\end{proposition}

\begin{enumerate}
\item $\left( F,A\right) \bar{\cup}\left( F,A\right) =\left( F,A\right) $

\item $\left( F,A\right) \bar{\cap}\left( F,A\right) =\left( F,A\right) $

\item $\left( F,A\right) \bar{\cup}$ $\emptyset =\left( F,A\right) $,\ where 
$\emptyset $ is a null bipolar fuzzy soft set.

\item $\left( F,A\right) \bar{\cap}$ $\emptyset =\emptyset $, where $%
\emptyset $ is a null bipolar fuzzy soft set.
\end{enumerate}

\begin{proof}
$\left( 1\right) $. $\left( F,A\right) \bar{\cup}\left( F,A\right) =\left(
F,A\right) $.

A bipolar fuzzy soft set $\left( H,C\right) $ is union of two bipolar fuzzy
soft sets $\left( F,A\right) $ and $\left( F,A\right) $ which is%
\begin{equation}
(H,C)=\left( F,A\right) \bar{\cup}\left( F,A\right) \text{ where }C=A\cup A
\label{70}
\end{equation}%
Define by%
\begin{eqnarray*}
H\left( e\right) &=&F\left( e\right) \text{ if }e\in A\backslash A \\
&=&F\left( e\right) \text{ if }e\in A\backslash A \\
&=&F\left( e\right) \cup F\left( e\right) \text{ if }e\in A\cap A
\end{eqnarray*}%
L.H.S. There are three cases.

Case (1).If $a\in A\backslash A$.%
\begin{equation*}
H\left( a\right) =F\left( a\right) \text{ if }a\in A\backslash A=\emptyset 
\end{equation*}%
Case (2) If $a\in A\backslash A$.%
\begin{equation*}
H\left( a\right) =F\left( a\right) \text{ if }a\in A\backslash A=\emptyset 
\end{equation*}%
Case (3) If $a\in A\cap A$.%
\begin{eqnarray*}
H\left( a\right)  &=&F\left( a\right) \cup F\left( a\right) \text{ if }a\in
A\cap A=A \\
&=&F\left( a\right) \text{ if }a\in A \\
H\left( a\right)  &=&F\left( a\right) \text{ if }a\in A \\
(H,C) &=&\left( F,A\right) \text{ from equation \ref{70}} \\
\left( F,A\right) \bar{\cup}\left( F,A\right)  &=&\left( F,A\right) \text{
from equation \ref{70}}
\end{eqnarray*}%
It is satisfied in all three cases. Hence $\left( F,A\right) \bar{\cup}%
\left( F,A\right) =\left( F,A\right) $.

2: $\left( F,A\right) \bar{\cap}\left( F,A\right) =\left( F,A\right) $

A bipolar fuzzy soft set $\left( H,C\right) $ is intersection of two bipolar
fuzzy soft sets $\left( F,A\right) $ and $\left( F,A\right) $ which is%
\begin{equation}
(H,C)=\left( F,A\right) \bar{\cap}\left( F,A\right) \text{ where }C=A\cap A
\label{71}
\end{equation}%
Define by%
\begin{equation*}
H\left( e\right) =F\left( e\right) \cap F\left( e\right) \text{ if }e\in
C=A\cap A
\end{equation*}%
L.H.S. Let a$\in $C=$A\cap A$.%
\begin{eqnarray*}
H\left( e\right) &=&F\left( e\right) \cap F\left( e\right) \text{ if }e\in
C=A\cap A=A \\
&=&F\left( e\right) \text{ if }e\in C=A \\
&=&F\left( e\right) \text{ if }e\in A \\
H\left( e\right) &=&F\left( e\right) \\
(H,C) &=&\left( F,A\right) \text{ from equation \ref{71}} \\
\left( F,A\right) \bar{\cap}\left( F,A\right) &=&\left( F,A\right) \text{
from equation \ref{71}}
\end{eqnarray*}%
Hence $\left( F,A\right) \bar{\cap}\left( F,A\right) =\left( F,A\right) $.
\end{proof}

\begin{lemma}
Absorption property of bipolar fuzzy soft sets $\left( F,A\right) $ and $%
\left( G,B\right) $.
\end{lemma}

\begin{enumerate}
\item $\left( F,A\right) \bar{\cup}\left( \left( F,A\right) \bar{\cap}\left(
G,B\right) \right) =\left( F,A\right) $

\item $\left( F,A\right) \bar{\cap}\left( \left( F,A\right) \bar{\cup}\left(
G,B\right) \right) =\left( F,A\right) $
\end{enumerate}

\begin{proof}
$\left( 1\right) $. $\left( F,A\right) \bar{\cup}\left( \left( F,A\right) 
\bar{\cap}\left( G,B\right) \right) =$ $\left( F,A\right) $

Let bipolar fuzzy soft set $\left( H,C\right) $ is an intersection of two
bipolar fuzzy soft sets $\left( F,A\right) $ and $\left( G,B\right) ,$ where 
$C=A\cap B$%
\begin{equation}
(H,C)=\left( F,A\right) \bar{\cap}\left( G,B\right) \text{ where }C=A\cap B
\label{7}
\end{equation}%
Define by%
\begin{equation*}
H\left( e\right) =F\left( e\right) \cap G\left( e\right) \text{ if }e\in
C=A\cap B
\end{equation*}%
Let bipolar fuzzy soft set $\left( K,M\right) $ is union of two bipolar
fuzzy soft sets $\left( F,A\right) $ and $\left( H,C\right) $ which is%
\begin{equation}
\left( K,M\right) =\left( F,A\right) \bar{\cup}(H,C)\text{ where }M=A\cup C
\label{72}
\end{equation}%
Define by%
\begin{eqnarray*}
K\left( e\right) &=&F\left( e\right) \text{ if }e\in A\backslash C \\
&=&H\left( e\right) \text{ if }e\in C\backslash A \\
&=&F\left( e\right) \cup H\left( e\right) \text{ if }e\in A\cap C
\end{eqnarray*}%
$L.H.S$. There are three cases.

Cases (1). If $e\in A\backslash C$.%
\begin{eqnarray*}
K\left( e\right) &=&F\left( e\right) \text{ if }e\in A\backslash C \\
&=&F\left( e\right) \text{ if }e\in A \\
K\left( e\right) &=&F\left( e\right) \\
\left( K,M\right) &=&\left( F,A\right) \text{ from equation \ref{72}}
\end{eqnarray*}

Case (2). If $e\in C\backslash A=A\cap B-A=0$.%
\begin{eqnarray*}
K\left( e\right) &=&H\left( e\right) \text{ if }e\in C\backslash A=0 \\
&=&\emptyset \text{ if }e\in \emptyset \\
K\left( e\right) &=&\emptyset \text{ if }e\in \emptyset \\
\left( K,M\right) &=&\emptyset \text{ from equation \ref{72}}
\end{eqnarray*}

Case (3). If $e\in A\cap C$.%
\begin{eqnarray*}
K\left( e\right) &=&F\left( e\right) \cup H\left( e\right) \text{ if }e\in
A\cap C\text{ and }C=A\cap B \\
&=&F\left( e\right) \cup \left( F\left( e\right) \cap G\left( e\right)
\right) \text{ from equation \ref{7}} \\
&=&F\left( e\right) \text{ since }\left( F\left( e\right) \cap G\left(
e\right) \right) \subset F\left( e\right) \\
&=&F\left( e\right) \\
K\left( e\right) &=&F\left( e\right) \\
\left( K,M\right) &=&\left( F,A\right) \text{ from equation \ref{72}}
\end{eqnarray*}%
It is satisfied in three cases. Hence $\left( F,A\right) \bar{\cup}\left(
\left( F,A\right) \bar{\cap}\left( G,B\right) \right) =$ $\left( F,A\right) $%
.

$\left( 2\right) $. same as above.
\end{proof}

\begin{theorem}
Commutative property of bipolar fuzzy soft sets $\left( F,A\right) $ and $%
\left( G,B\right) $.
\end{theorem}

$\left( 1\right) $ $\left( F,A\right) \bar{\cap}\left( G,B\right) =\left(
G,B\right) \bar{\cap}\left( F,A\right) $

$\left( 2\right) $ $\left( F,A\right) \bar{\cup}\left( G,B\right) =\left(
G,B\right) \bar{\cup}\left( F,A\right) $

\begin{proof}
$\left( 1\right) .$ To show that $\left( F,A\right) \bar{\cap}\left(
G,B\right) =\left( G,B\right) \bar{\cap}\left( F,A\right) $.

A bipolar fuzzy soft set $\left( H,C\right) $ is an intersection of two
bipolar fuzzy soft sets $\left( F,A\right) $ and $\left( G,B\right) ,$ where 
$C=A\cap B$%
\begin{equation}
H\left( e\right) =F\left( e\right) \cap G\left( e\right) \text{ if }e\in
C=A\cap B  \label{5}
\end{equation}

A bipolar fuzzy soft set $\left( K,D\right) $ is an intersection of two
bipolar fuzzy soft sets $\left( G,B\right) $ and $\left( F,A\right) $, where 
$D=B\cap A$%
\begin{equation}
K\left( e\right) =G\left( e\right) \cap F\left( e\right) \text{ if }e\in
D=B\cap A  \label{6}
\end{equation}%
To show that $\left( H,C\right) =\left( K,D\right) $

L.H.S%
\begin{eqnarray*}
H\left( e\right) &=&F\left( e\right) \cap G\left( e\right) \text{ for all }%
e\in C=A\cap B \\
&=&G\left( e\right) \cap F\left( e\right) \text{ since }F\left( e\right)
\cap G\left( e\right) =G\left( e\right) \cap F\left( e\right) \\
&=&G\left( e\right) \cap F\left( e\right) \text{ for all }e\in C=A\cap
B=B\cap A=D \\
&=&K\left( e\right) \text{ for all }e\in B\cap A=D \\
H\left( e\right) &=&K\left( e\right) \\
\left( H,C\right) &=&\left( K,D\right) \text{ from equation \ref{5},
equation \ref{6}} \\
\left( F,A\right) \bar{\cap}\left( G,B\right) &=&\left( G,B\right) \bar{\cap}%
\left( F,A\right) \text{ from equation \ref{5}, equation \ref{6}}
\end{eqnarray*}%
Hence $\left( F,A\right) \bar{\cap}\left( G,B\right) =\left( G,B\right) \bar{%
\cap}\left( F,A\right) $

$\left( 2\right) .$ To show that $\left( F,A\right) \bar{\cup}\left(
G,B\right) =\left( G,B\right) \bar{\cup}\left( F,A\right) $.

L.H.S.

A bipolar fuzzy soft set\ $\left( H,C\right) $ is union of two bipolar fuzzy
soft sets $\left( F,A\right) $, $\left( G,B\right) $ over a common universe $%
U$%
\begin{equation}
\left( H,C\right) =\left( F,A\right) \bar{\cup}\left( G,B\right) \text{
where }C=A\cup B  \label{83}
\end{equation}%
Define by%
\begin{eqnarray}
H\left( e\right) &=&F\left( e\right) \text{ if }e\in A\backslash B
\label{87} \\
&=&G\left( e\right) \text{ if }e\in B\backslash A  \label{88} \\
&=&F\left( e\right) \cup G\left( e\right) \text{ if }e\in A\cap B  \label{89}
\end{eqnarray}%
There are three cases.

Case (1). If $e\in A\backslash B$%
\begin{equation}
H\left( e\right) =F\left( e\right) \text{ if }e\in A\backslash B\text{ from
equation \ref{87}}  \label{84}
\end{equation}

Case (2). If $e\in B\backslash A$%
\begin{equation}
H\left( e\right) =G\left( e\right) \text{ if }e\in B\backslash A\text{ from
equation \ref{88}}  \label{85}
\end{equation}

Case (3). If $e\in A\cap B$%
\begin{eqnarray}
H\left( e\right) &=&F\left( e\right) \cup G\left( e\right) \text{ if }e\in
A\cap B=B\cap A\text{ from equation \ref{89}}  \notag \\
&=&G\left( e\right) \cup F\left( e\right) \text{ if }e\in B\cap A  \label{86}
\end{eqnarray}

Combine equation \ref{84}, equation \ref{85} and equation \ref{86}. We get%
\begin{eqnarray*}
H\left( e\right) &=&G\left( e\right) \text{ if }e\in B\backslash A \\
&=&F\left( e\right) \text{ if }e\in A\backslash B \\
&=&G\left( e\right) \cup F\left( e\right) \text{ if }e\in B\cap A
\end{eqnarray*}

$(H,C)$ becomes%
\begin{eqnarray*}
\left( H,C\right) &=&\left( G,B\right) \bar{\cup}\left( F,A\right) \text{
where }C=B\cup A \\
&=&R.H.S
\end{eqnarray*}

Hence $\left( F,A\right) \bar{\cup}\left( G,B\right) =\left( G,B\right) \bar{%
\cup}\left( F,A\right) $.
\end{proof}

\begin{theorem}
Associative law of bipolar fuzzy soft sets $\left( F,A\right) $,$\left(
G,B\right) $ and $\left( H,C\right) $.
\end{theorem}

$\left( \text{\textexclamdown }\right) $ $\left( F,A\right) \bar{\cap}\left(
\left( G,B\right) \bar{\cap}\left( H,C\right) \right) =\left( \left(
F,A\right) \bar{\cap}\left( G,B\right) \right) \bar{\cap}\left( H,C\right) $

$\left( \text{\textexclamdown \textexclamdown }\right) $ $\left( F,A\right) 
\bar{\cup}\left( \left( G,B\right) \bar{\cup}\left( H,C\right) \right)
=\left( \left( F,A\right) \bar{\cup}\left( G,B\right) \right) \bar{\cup}%
\left( H,C\right) $

\begin{proof}
$\left( \text{\textexclamdown }\right) $ $\left( F,A\right) \bar{\cap}\left(
\left( G,B\right) \bar{\cap}\left( H,C\right) \right) =\left( \left(
F,A\right) \bar{\cap}\left( G,B\right) \right) \bar{\cap}\left( H,C\right) $.

A bipolar fuzzy soft set $\left( L,D\right) $ is an intersection of two
bipolar fuzzy soft sets $\left( G,B\right) $ and $\left( H,C\right) $ which
is%
\begin{equation}
\left( G,B\right) \bar{\cap}\left( H,C\right) =(L,D)\text{ where }D=B\cap C
\label{1}
\end{equation}%
Define by%
\begin{equation}
L\left( e\right) =G\left( e\right) \cap H\left( e\right) \text{ if }e\in
D=B\cap C  \label{74}
\end{equation}%
A bipolar fuzzy soft set $\left( M,X\right) $ is an intersection of two
bipolar fuzzy soft sets $\left( F,A\right) $ and $\left( L,D\right) $ which
is%
\begin{equation}
\left( F,A\right) \bar{\cap}(L,D)=\left( M,X\right) \text{ where }X=A\cap D
\label{73}
\end{equation}%
Define by%
\begin{equation}
M\left( e\right) =F\left( e\right) \cap L\left( e\right) \text{ if }e\in
X=A\cap D  \label{75}
\end{equation}%
L.H.S:%
\begin{eqnarray*}
M\left( e\right) &=&F\left( e\right) \cap L\left( e\right) \text{ for all }%
e\in X=A\cap D\text{ from equation \ref{75}} \\
&=&F\left( e\right) \cap \left( G\left( e\right) \cap H\left( e\right)
\right) \text{ from equation \ref{74}} \\
&=&\left( F\left( e\right) \cap G\left( e\right) \right) \cap H\left(
e\right) \text{ for all }e\in X=A\cap D=A\cap \left( B\cap C\right) \\
M\left( e\right) &=&\left( F\left( e\right) \cap G\left( e\right) \right)
\cap H\left( e\right) \text{ for all }e\in A\cap \left( B\cap C\right) \\
\left( M,X\right) &=&\left( \left( F,A\right) \bar{\cap}\left( G,B\right)
\right) \bar{\cap}\left( H,C\right) \text{ from equation \ref{73}} \\
\left( F,A\right) \bar{\cap}(L,D) &=&\left( \left( F,A\right) \bar{\cap}%
\left( G,B\right) \right) \bar{\cap}\left( H,C\right) \text{ from equation %
\ref{73}} \\
\left( F,A\right) \bar{\cap}\left( \left( G,B\right) \bar{\cap}\left(
H,C\right) \right) &=&\left( \left( F,A\right) \bar{\cap}\left( G,B\right)
\right) \bar{\cap}\left( H,C\right) \text{ from equation \ref{1}}
\end{eqnarray*}%
Hence $\left( F,A\right) \bar{\cap}\left( \left( G,B\right) \bar{\cap}\left(
H,C\right) \right) =\left( \left( F,A\right) \bar{\cap}\left( G,B\right)
\right) \bar{\cap}\left( H,C\right) $.

$\left( \mathbf{ii}\right) $.\textbf{\ }same as above.

Hence $\left( F,A\right) \bar{\cup}\left( \left( G,B\right) \bar{\cup}\left(
H,C\right) \right) =\left( \left( F,A\right) \bar{\cup}\left( G,B\right)
\right) \bar{\cup}\left( H,C\right) $.
\end{proof}

\begin{theorem}
. Distributive law of bipolar fuzzy soft sets $\left( F,A\right) $, $\left(
G,B\right) $ and $\left( H,C\right) $.
\end{theorem}

\begin{enumerate}
\item $\left( F,A\right) \bar{\cap}\left( \left( G,B\right) \bar{\cup}\left(
H,C\right) \right) =\left( \left( F,A\right) \bar{\cap}\left( G,B\right)
\right) \bar{\cup}\left( \left( F,A\right) \bar{\cap}\left( H,C\right)
\right) $

\item $\left( F,A\right) \bar{\cup}\left( \left( G,B\right) \bar{\cap}\left(
H,C\right) \right) =\left( \left( F,A\right) \bar{\cup}\left( G,B\right)
\right) \bar{\cap}\left( \left( F,A\right) \bar{\cup}\left( H,C\right)
\right) $
\end{enumerate}

\begin{proof}
$(1):$ $\left( F,A\right) \bar{\cap}\left( \left( G,B\right) \bar{\cup}%
\left( H,C\right) \right) =\left( \left( F,A\right) \bar{\cap}\left(
G,B\right) \right) \bar{\cup}\left( \left( F,A\right) \bar{\cap}\left(
H,C\right) \right) $

A bipolar fuzzy soft set\ $\left( L,D\right) $ is union of two bipolar fuzzy
soft sets $\left( G,B\right) $ and $\left( H,C\right) $ over a common
universe $U$.

\begin{equation}
\left( G,B\right) \bar{\cup}\left( H,C\right) =(L,D)\text{ where }D=B\cup C
\label{2}
\end{equation}%
Define by%
\begin{eqnarray*}
L\left( e\right) &=&G\left( e\right) \text{ if }e\in B\backslash C \\
&=&H\left( e\right) \text{ if }e\in C\backslash B \\
&=&G\left( e\right) \cup H\left( e\right) \text{ if }e\in B\cap C
\end{eqnarray*}%
A bipolar fuzzy soft set $\left( M,V\right) $ is an intersection of two
bipolar fuzzy soft sets $\left( F,A\right) $ and $\left( L,D\right) $.%
\begin{equation}
\left( F,A\right) \bar{\cap}\left( L,D\right) =(M,V)\text{ where }V=A\cap D
\label{3}
\end{equation}%
Define by%
\begin{equation}
M\left( e\right) =F\left( e\right) \cap L\left( e\right) \text{ if }e\in
V=A\cap D  \label{78}
\end{equation}%
L.H.S%
\begin{eqnarray}
M\left( e\right) &=&F\left( e\right) \cap L\left( e\right) \text{ for all }%
e\in V=A\cap D\text{ from equation \ref{78}}  \notag \\
M\left( e\right) &=&F\left( e\right) \cap L\left( e\right) \text{ for all }%
e\in A\cap D\text{ so }e\in A\text{, }e\in D  \label{79}
\end{eqnarray}%
If $e\in D=B\cup C$ from equation \ref{2}. Then there are three cases.

Case (1) If $e\in B\backslash C$%
\begin{equation}
L\left( e\right) =G\left( e\right) \text{ if }e\in B\backslash C\text{ from
equation \ref{2}}  \label{80}
\end{equation}%
Case (2) \ If $e\in C\backslash B$%
\begin{equation}
L\left( e\right) =H\left( e\right) \text{ if }e\in C\backslash B\text{ from
equation \ref{2}}  \label{81}
\end{equation}%
Case (3) \ If $e\in C\cap B$%
\begin{equation}
L\left( e\right) =G\left( e\right) \cup H\left( e\right) \text{ if }e\in
B\cap C\text{ from equation \ref{2}}  \label{82}
\end{equation}%
Put equation \ref{80},equation \ref{81} and equation \ref{82} in equation %
\ref{79}%
\begin{eqnarray*}
M\left( e\right) &=&F\left( e\right) \cap G(e)\text{ for all }e\in A\text{, }%
e\in B\backslash C \\
&=&F\left( e\right) \cap H\left( e\right) \text{ if }e\in A\text{, }e\in
C\backslash B \\
&=&F\left( e\right) \cap \left( G\left( e\right) \cup H\left( e\right)
\right) \text{ if }e\in A\text{, }e\in B\cap C \\
&=&\left( F\left( e\right) \cap G\left( e\right) \right) \cup \left( F\left(
e\right) \cap H\left( e\right) \right) \\
M\left( e\right) &=&\left( F\left( e\right) \cap G\left( e\right) \right)
\cup \left( F\left( e\right) \cap H\left( e\right) \right) \\
(M,V) &=&\left( \left( F,A\right) \bar{\cap}\left( G,B\right) \right) \bar{%
\cup}\left( \left( F,A\right) \bar{\cap}\left( H,C\right) \right) \text{
from equation \ref{3}} \\
\left( F,A\right) \bar{\cap}\left( L,D\right) &=&\left( \left( F,A\right) 
\bar{\cap}\left( G,B\right) \right) \bar{\cup}\left( \left( F,A\right) \bar{%
\cap}\left( H,C\right) \right) \text{ from equation \ref{3}} \\
\left( F,A\right) \bar{\cap}\left( \left( G,B\right) \bar{\cup}\left(
H,C\right) \right) &=&\left( \left( F,A\right) \bar{\cap}\left( G,B\right)
\right) \bar{\cup}\left( \left( F,A\right) \bar{\cap}\left( H,C\right)
\right) \text{ from equation \ref{2}}
\end{eqnarray*}%
Hence $\left( F,A\right) \bar{\cap}\left( \left( G,B\right) \bar{\cup}\left(
H,C\right) \right) =\left( \left( F,A\right) \bar{\cap}\left( G,B\right)
\right) \bar{\cup}\left( \left( F,A\right) \bar{\cap}\left( H,C\right)
\right) $.

$(2)$. same as above.
\end{proof}

\begin{lemma}
: $\left( F,A\right) $ and $\left( G,B\right) $ are two bipolar fuzzy soft
sets.
\end{lemma}

\begin{enumerate}
\item $\left( F,A\right) \subset \left( G,B\right) \Rightarrow \left(
F,A\right) \bar{\cap}\left( G,B\right) =\left( F,A\right) $

\item $\left( F,A\right) \subset \left( G,B\right) \Rightarrow \left(
F,A\right) \bar{\cup}\left( G,B\right) =\left( G,B\right) $.
\end{enumerate}

\section{De Morgan's Law of Bipolar Fuzzy Soft Sets}

\begin{theorem}
. De Morgan's law of bipolar fuzzy soft sets $\left( F,A\right) $ and $%
\left( G,B\right) $.
\end{theorem}

\begin{enumerate}
\item $\left( \left( F,A\right) \bar{\cup}\left( G,B\right) \right)
^{c}=\left( F,A\right) ^{c}\sqcap _{\epsilon }\left( G,B\right) ^{c}$.

\item $\left( \left( F,A\right) \sqcap _{\epsilon }\left( G,B\right) \right)
^{c}=\left( F,A\right) ^{c}\bar{\cup}\left( G,B\right) ^{c}$.
\end{enumerate}

\begin{proof}
$\left( 1\right) $.

Let $\left( F,A\right) $ and $\left( G,B\right) $ be a two bipolar fuzzy
soft sets over a common universe $U$. Then the union of two bipolar fuzzy
soft sets $\left( F,A\right) $ and $\left( G,B\right) $ is a bipolar fuzzy
soft set $\left( H\text{,}C\right) $ where $C=A\cup B$ and $\left( H\text{,}%
C\right) =\left( F,A\right) \bar{\cup}\left( G,B\right) $ is define by%
\begin{eqnarray}
H\left( e\right) &=&F\left( e\right) \text{ if }e\in A\backslash B
\label{29} \\
&=&G\left( e\right) \text{ if }e\in B\backslash A  \notag \\
&=&F\left( e\right) \cup G\left( e\right) \text{ if }e\in A\cap B  \notag
\end{eqnarray}%
The extended intersection of two bipolar fuzzy soft sets $\left( F,A\right) $
and $\left( G,B\right) $ is bipolar fuzzy soft set $\left( K\text{,}D\right) 
$ where $D=A\cap B$ and $\left( K\text{,}D\right) =\left( F,A\right) \sqcap
_{\epsilon }\left( G,B\right) $ is define by%
\begin{eqnarray}
K\left( e\right) &=&F\left( e\right) \text{ if }e\in A\backslash B
\label{30} \\
&=&G\left( e\right) \text{ if }e\in B\backslash A  \notag \\
&=&F\left( e\right) \cap G\left( e\right) \text{ if }e\in A\cap B  \notag
\end{eqnarray}%
To show that $\left( \left( F,A\right) \bar{\cup}\left( G,B\right) \right)
^{c}=\left( F,A\right) ^{c}\sqcap _{\epsilon }\left( G,B\right) ^{c}$.

L.H.S: There are three cases.

Case (i): If $e\in A\backslash B$. Then $e\in C$. Such that%
\begin{equation*}
H\left( e\right) =F\left( e\right) \text{ if }e\in A\backslash B\text{ from
equation \ref{29}}
\end{equation*}

Taking complement of above. So%
\begin{equation}
\left( H\left( e\right) \right) ^{c}=\left( F\left( e\right) \right) ^{c}%
\text{ if }e\in A\backslash B  \label{103}
\end{equation}

Case (ii) If $e\in B\backslash A$. Then $e\in C$. Such that 
\begin{equation*}
H\left( e\right) =G\left( e\right) \text{ if }e\in B\backslash A\text{ from %
\ref{29}}
\end{equation*}

Taking complement of above. So%
\begin{equation}
\left( H\left( e\right) \right) ^{c}=\left( G\left( e\right) \right) ^{c}%
\text{ if }e\in B\backslash A  \label{108}
\end{equation}

Case (iii) If $e\in A\cap B$. Then $e\in C$. Such that%
\begin{equation*}
H\left( e\right) =F\left( e\right) \cup G\left( e\right) \text{ if }e\in
A\cap B\text{ from \ref{29}}
\end{equation*}

Taking complement of above. So%
\begin{equation}
\left( H\left( e\right) \right) ^{c}=\left( F\left( e\right) \cup G\left(
e\right) \right) ^{c}\text{ if }e\in A\cap B  \label{34}
\end{equation}

We define $F\left( e\right) $ and $G\left( e\right) $ as%
\begin{equation}
F\left( e\right) =\text{\{(}u\text{,}%
{\mu}%
_{A}^{+}\left( u\right) \text{,}%
{\mu}%
_{A}^{-}\left( u\right) \text{) :}u\in U\text{\}}  \label{35}
\end{equation}

and%
\begin{equation}
G\left( e\right) =\text{\{(}u\text{,}%
{\mu}%
_{B}^{+}\left( u\right) \text{,}%
{\mu}%
_{B}^{-}\left( u\right) \text{) :}u\in U\text{\}}  \label{36}
\end{equation}

Putting equation \ref{35} , equation \ref{36} in equation \ref{34}. We get%
\begin{eqnarray}
\left( H\left( e\right) \right) ^{c} &=&\left( \text{(}u\text{,}%
{\mu}%
_{A}^{+}\left( u\right) \text{,}%
{\mu}%
_{A}^{-}\left( u\right) \text{)}\cup \text{(}u\text{,}%
{\mu}%
_{B}^{+}\left( u\right) \text{,}%
{\mu}%
_{B}^{-}\left( u\right) \text{)}\right) ^{c}\text{ if }e\in A\cap B  \notag
\\
&=&\text{(}u\text{,max}\left( 
{\mu}%
_{A}^{+}\left( u\right) \text{,}%
{\mu}%
_{B}^{+}\left( u\right) \right) \text{,min}\left( 
{\mu}%
_{A}^{-}\left( u\right) \text{,}%
{\mu}%
_{B}^{-}\left( u\right) \right) \text{)}^{c}  \notag \\
&=&\text{(}u\text{,}\left( 1-\text{max}\left( 
{\mu}%
_{A}^{+}\left( u\right) \text{,}%
{\mu}%
_{B}^{+}\left( u\right) \right) \right) \text{,}\left( -1-\left( -\text{min}%
\left( 
{\mu}%
_{A}^{-}\left( u\right) \text{,}%
{\mu}%
_{B}^{-}\left( u\right) \right) \right) \right) \text{)}  \notag \\
&=&\text{(}u\text{,min}\left( 1-%
{\mu}%
_{A}^{+}\left( u\right) \text{,}1-%
{\mu}%
_{B}^{+}\left( u\right) \right) \text{,max}\left( -1-\left( -%
{\mu}%
_{A}^{-}\left( u\right) \right) \text{,}-1-\left( -%
{\mu}%
_{B}^{-}\left( u\right) \right) \right) \text{)}  \notag \\
&=&\text{(}u\text{,}1-%
{\mu}%
_{A}^{+}\left( u\right) \text{,}-1-\left( -%
{\mu}%
_{A}^{-}\left( u\right) \right) \text{)}\cap \text{(}u\text{,}1-%
{\mu}%
_{B}^{+}\left( u\right) \text{,}-1-\left( -%
{\mu}%
_{B}^{-}\left( u\right) \right) \text{)}  \notag \\
&=&\text{(}u\text{,}%
{\mu}%
_{A}^{+}\left( u\right) \text{,}%
{\mu}%
_{A}^{-}\left( u\right) \text{)}^{c}\cap \text{(}u\text{,}%
{\mu}%
_{B}^{+}\left( u\right) \text{,}%
{\mu}%
_{B}^{-}\left( u\right) \text{)}^{c}\text{ if }e\in A\cap B  \notag \\
&=&\left( F\left( e\right) \right) ^{c}\cap \left( G\left( e\right) \right)
^{c}\text{ if }e\in A\cap B\text{ and }C=A\cap B  \label{39}
\end{eqnarray}

From equation \ref{103}, equation \ref{108} and equation \ref{39}. We get%
\begin{eqnarray*}
\left( H\left( e\right) \right) ^{c} &=&\left( F\left( e\right) \right) ^{c}%
\text{ if }e\in A\backslash B \\
&=&\left( G\left( e\right) \right) ^{c}\text{ if }e\in B\backslash A \\
&=&\left( F\left( e\right) \right) ^{c}\cap \left( G\left( e\right) \right)
^{c}\text{ if }e\in A\cap B\text{ and }C=A\cap B
\end{eqnarray*}

Then

\begin{equation*}
\left( H\left( e\right) \right) ^{c}=\left( F\left( e\right) \right)
^{c}\sqcap _{\epsilon }\left( G\left( e\right) \right) ^{c}
\end{equation*}

Thus%
\begin{equation*}
\left( \left( F,A\right) \bar{\cup}\left( G,B\right) \right) ^{c}=\left(
F,A\right) ^{c}\sqcap _{\epsilon }\left( G,B\right) ^{c}
\end{equation*}

Hence it is proved.

(2). same as above.
\end{proof}

\section{OR and AND Operations on Bipolar Fuzzy Soft Sets}

\begin{definition}
. Let $\left( F,A\right) $ and $\left( G,B\right) $ be two bipolar fuzzy
soft sets over a common universe $U$. Then,
\end{definition}

\begin{enumerate}
\item $\left( F,A\right) \tilde{\wedge}\left( G,B\right) $ is a bipolar
fuzzy soft set defined by $\left( F,A\right) \tilde{\wedge}\left( G,B\right)
=\left( H,A\times B\right) $ where $H\left( a,b\right) =F\left( a\right)
\cap G\left( b\right) $ for all $\left( a,b\right) ,\in C=A\times B$, where $%
\cap $ is the intersection operation of sets.

\item $\left( F,A\right) \tilde{\vee}\left( G,B\right) $ is a bipolar fuzzy
soft set defined by $\left( F,A\right) \tilde{\vee}\left( G,B\right) =\left(
H,A\times B\right) $ where $H\left( a,b\right) =F\left( a\right) \cup
G\left( b\right) $ for all $\left( a,b\right) \in C=A\times B$, where $\cup $
is the intersection operation of sets.
\end{enumerate}

\begin{definition}
. Let $T=$\{$\left( F_{i},A_{i}\right) $ : $i\in I$\} be a family of bipolar
fuzzy soft sets in a bipolar fuzzy soft class $(U,E)$. Then the intersection
of bipolar fuzzy soft sets in $T$ is bipolar fuzzy soft set $\left(
H,C\right) $, where, $C=\times A_{i}$ for all $i\in I$, $H\left( e\right) =%
\tilde{\wedge}F_{i}\left( e\right) $ for all $e\in C$, $\left( H,C\right) =%
\tilde{\wedge}\left( F_{i},A_{i}\right) $ for all $i\in I$.
\end{definition}

\begin{definition}
. Let $T=$\{$\left( F_{i},A_{i}\right) $ : $i\in I$\} be a family of bipolar
fuzzy soft sets in a bipolar fuzzy soft class $(U,E)$. Then the union of
bipolar fuzzy soft sets in $T$ is bipolar fuzzy soft set $\left( H,C\right) $%
, where, $C=\times A_{i}$ for all $i\in I$,$H\left( e\right) =\tilde{\vee}%
F_{i}\left( e\right) $ for all $e\in C$, $\left( H,C\right) =\tilde{\vee}%
\left( F_{i},A_{i}\right) $ for all $i\in I$.
\end{definition}

\begin{example}
. Let $U=$\{$m_{1}$, $m_{2}$, $m_{3}$, $m_{4}$\} be the set of four man
under consideration and $E=$\{$e_{1}=$Educated, $e_{2}=$Government employee, 
$e_{3}=$Businessman, $e_{4}=$Smart, $e_{5}=$Weak\} be the set of parameters
and $A=$\{$e_{1}$, $e_{2}$\}$\subseteq E$, $B=$\{$e_{4}$, $e_{5}$\}$%
\subseteq E$. Then%
\begin{equation*}
\left( F,A\right) =\left\{ 
\begin{array}{c}
F\left( e_{1}\right) =\left\{ 
\begin{array}{c}
\left( m_{1},0.1,-0.5\right) ,\left( m_{2},0.3,-0.6\right) \\ 
\left( m_{3},0.4,-0.2\right) ,\left( m_{4},0.7,-0.2\right)%
\end{array}%
\right\} , \\ 
F\left( e_{2}\right) =\left\{ 
\begin{array}{c}
\left( m_{1},0.3,-0.5\right) ,\left( m_{2},0.4,-0.2\right) , \\ 
\left( m_{3},0.5,-0.2\right) ,\left( m_{4},0.4,-0.2\right)%
\end{array}%
\right\}%
\end{array}%
\right\}
\end{equation*}%
\begin{equation*}
\left( G,B\right) =\left\{ 
\begin{array}{c}
G\left( e_{4}\right) =\left\{ 
\begin{array}{c}
\left( m_{1},0.1,-0.6\right) ,\left( m_{2},0.3,-0.4\right) , \\ 
\left( m_{3},0.1,-0.6\right) ,\left( m_{4},0.0,-0.2\right)%
\end{array}%
\right\} , \\ 
G\left( e_{5}\right) =\left\{ 
\begin{array}{c}
\left( m_{1},0.4,-0.1\right) ,\left( m_{2},0.2,-0.4\right) , \\ 
\left( m_{3},0.6,-0.4\right) ,\left( m_{4},0.7,-0.0\right)%
\end{array}%
\right\}%
\end{array}%
\right\}
\end{equation*}%
Then $\left( F,A\right) \tilde{\wedge}\left( G,B\right) =\left( H,A\times
B\right) $ and $C=A\times B=$\{$e_{1}$, $e_{2}$\}$\times $\{$e_{4}$, $e_{5}$%
\}$=$\{$\left( e_{1}\text{, }e_{4}\right) $, $\left( e_{1}\text{, }%
e_{5}\right) $, $\left( e_{2}\text{, }e_{4}\right) $, $\left( e_{2}\text{, }%
e_{5}\right) $\}\ define by $H\left( a,b\right) =F\left( a\right) \cap
G\left( b\right) $ for all $\left( a,b\right) \in C$ 
\begin{equation*}
\left( H,C\right) =\left( F,A\right) \tilde{\wedge}\left( G,B\right)
=\left\{ 
\begin{array}{c}
H\left( e_{1},e_{4}\right) =\left\{ 
\begin{array}{c}
\left( m_{1},0.1,-0.5\right) ,\left( m_{2},0.3,-0.4\right) , \\ 
\left( m_{3},0.1,-0.2\right) ,\left( m_{4},0.0,-0.2\right)%
\end{array}%
\right\} , \\ 
H\left( e_{1},e_{5}\right) =\left\{ 
\begin{array}{c}
\left( m_{1},0.1,-0.1\right) ,\left( m_{2},0.2,-0.4\right) , \\ 
\left( m_{3},0.4,-0.2\right) ,\left( m_{4},0.7,-0.0\right)%
\end{array}%
\right\} , \\ 
H\left( e_{2},e_{4}\right) =\left\{ 
\begin{array}{c}
\left( m_{1},0.1,-0.5\right) ,\left( m_{2},0.3,-0.2\right) , \\ 
\left( m_{3},0.1,-0.2\right) ,\left( m_{4},0.0,-0.2\right)%
\end{array}%
\right\} , \\ 
H\left( e_{2},e_{5}\right) =\left\{ 
\begin{array}{c}
\left( m_{1},0.3,-0.1\right) ,\left( m_{2},0.2,-0.2\right) , \\ 
\left( m_{3},0.5,-0.2\right) ,\left( m_{4},0.4,-0.0\right)%
\end{array}%
\right\}%
\end{array}%
\right\}
\end{equation*}
\end{example}

\begin{example}
Let $U=$\{$h_{1}$, $h_{2}$, $h_{3}$, $h_{4}$\} be the set of four houses
under consideration and $E=$\{$e_{1}$=Expensive, $e_{2}$=Beautiful, $e_{3}$%
=Wooden, $e_{4}$=In the green surrounding, $e_{5}$=Convenient traffic\} be
the set of parameters and $A=$\{$e_{1}$, $e_{2}$\}$\subseteq E$, $B=$\{$e_{4}
$, $e_{5}$\}$\subseteq E$. Then,%
\begin{equation*}
\left( F,A\right) =\left\{ 
\begin{array}{c}
F\left( e_{1}\right) =\left\{ 
\begin{array}{c}
\left( h_{1},0.1,-0.5\right) ,\left( h_{2},0.3,-0.6\right) , \\ 
\left( h_{3},0.4,-0.2\right) ,\left( h_{4},0.7,-0.2\right) 
\end{array}%
\right\} , \\ 
F\left( e_{2}\right) =\left\{ 
\begin{array}{c}
\left( h_{1},0.3,-0.5\right) ,\left( h_{2},0.4,-0.2\right) , \\ 
\left( h_{3},0.5,-0.2\right) ,\left( h_{4},0.4,-0.2\right) 
\end{array}%
\right\} ,%
\end{array}%
\right\} 
\end{equation*}%
\begin{equation*}
\left( G,B\right) =\left\{ 
\begin{array}{c}
G\left( e_{4}\right) =\left\{ 
\begin{array}{c}
\left( h_{1},0.1,-0.6\right) ,\left( h_{2},0.3,-0.4\right) , \\ 
\left( h_{3},0.1,-0.6\right) ,\left( h_{4},0.0,-0.2\right) 
\end{array}%
\right\} , \\ 
G\left( e_{5}\right) =\left\{ 
\begin{array}{c}
\left( h_{1},0.4,-0.1\right) ,\left( h_{2},0.2,-0.4\right) , \\ 
\left( h_{3},0.6,-0.4\right) ,\left( h_{4},0.7,-0.0\right) 
\end{array}%
\right\} 
\end{array}%
\right\} 
\end{equation*}%
Then $\left( F,A\right) \tilde{\vee}\left( G,B\right) =\left( H,A\times
B\right) $ and $C=A\times B=$\{$e_{1}$, $e_{2}$\}$\times $\{$e_{4}$, $e_{5}$%
\}$=$\{$\left( e_{1}\text{, }e_{4}\right) $, $\left( e_{1}\text{, }%
e_{5}\right) $, $\left( e_{2}\text{, }e_{4}\right) $, $\left( e_{2}\text{, }%
e_{5}\right) $\}\ define by $H\left( a\right) =F\left( a\right) \cup G\left(
a\right) $, for all $a\in $ $C=A\times B$%
\begin{equation*}
\left( H,C\right) =\left( F,A\right) \tilde{\vee}\left( G,B\right) =\left\{ 
\begin{array}{c}
H\left( e_{1},e_{4}\right) =\left\{ 
\begin{array}{c}
\left( h_{1},0.1,-0.6\right) ,\left( h_{2},0.3,-0.6\right) , \\ 
\left( h_{3},0.4,-0.6\right) ,\left( h_{4},0.7,-0.2\right) 
\end{array}%
\right\} , \\ 
H\left( e_{1},e_{5}\right) =\left\{ 
\begin{array}{c}
\left( h_{1},0.4,-0.5\right) ,\left( h_{2},0.3,-0.6\right) , \\ 
\left( h_{3},0.6,-0.4\right) ,\left( h_{4},0.7,-0.2\right) 
\end{array}%
\right\} , \\ 
H\left( e_{2},e_{4}\right) =\left\{ 
\begin{array}{c}
\left( h_{1},0.3,-0.6\right) ,\left( h_{2},0.4,-0.4\right) , \\ 
\left( h_{3},0.5,-0.6\right) ,\left( h_{4},0.4,-0.2\right) 
\end{array}%
\right\} , \\ 
H\left( e_{2},e_{5}\right) =\left\{ 
\begin{array}{c}
\left( h_{1},0.4,-0.5\right) ,\left( h_{2},0.4,-0.4\right) , \\ 
\left( h_{3},0.6,-0.4\right) ,\left( h_{4},0.7,-0.2\right) 
\end{array}%
\right\} 
\end{array}%
\right\} 
\end{equation*}
\end{example}

\begin{proposition}
\textbf{Idempotent Property}. If $\left( F,A\right) ,\left( G,B\right) $ are
two bipolar fuzzy soft sets over $U$. Then,
\end{proposition}

\begin{enumerate}
\item $\left( F,A\right) \tilde{\wedge}\left( F,A\right) =\left( F,A\right) $

\item $\left( F,A\right) \tilde{\vee}\left( F,A\right) =\left( F,A\right) $
\end{enumerate}

\begin{proof}
$\left( 1\right) $.$\ \left( F,A\right) \tilde{\wedge}\left( F,A\right)
=\left( F,A\right) $

Suppose that $\left( F,A\right) \tilde{\wedge}\left( F,A\right) =\left(
H,C\right) $, where $C=A\times A$, Let $a\in A$%
\begin{eqnarray*}
H\left( a,a\right) &=&\text{ }F\left( a\right) \cap F\left( a\right) \text{
since }F\left( a\right) \cap F\left( a\right) =F\left( a\right) \\
&=&F\left( a\right) \\
H\left( a,a\right) &=&\text{ }F\left( a\right) \\
\left( H,C\right) &=&\left( F,A\right) \\
\left( F,A\right) \tilde{\wedge}\left( F,A\right) &=&\left( F,A\right)
\end{eqnarray*}%
Hence $\left( F,A\right) \tilde{\wedge}\left( F,A\right) =\left( F,A\right) $

$\left( 2\right) $. $\left( F,A\right) \tilde{\vee}\left( F,A\right) =\left(
F,A\right) $

Suppose that $\left( F,A\right) \tilde{\vee}\left( F,A\right) =\left(
H,C\right) $, where $C=A\times A$, Let $a\in A$%
\begin{eqnarray*}
H\left( a,a\right)  &=&\text{ }F\left( a\right) \cap F\left( a\right) \text{
since }F\left( a\right) \cup F\left( a\right) =F\left( a\right)  \\
&=&F\left( a\right)  \\
H\left( a,a\right)  &=&\text{ }F\left( a\right)  \\
\left( H,C\right)  &=&\left( F,A\right)  \\
\left( F,A\right) \tilde{\vee}\left( F,A\right)  &=&\left( F,A\right) 
\end{eqnarray*}%
Hence $\left( F,A\right) \tilde{\vee}\left( F,A\right) =\left( F,A\right) $.
\end{proof}

\begin{example}
Let $U=$\{$c_{1}$, $c_{2}$, $c_{3}$, $c_{4}$\} be the set of four cars under
consideration, $E=$\{$e_{1}$=Costly, $e_{2}$=Beautiful, $e_{3}$=Fuel
efficient,$e_{4}$=Luxurious\} be the set of parameters and $A=$\{$e_{1}$, $%
e_{2}$, $e_{3}$\}$\subset E$ then $F:A\rightarrow BF^{U}$ define by%
\begin{equation*}
F\left( e_{1}\right) =\left\{ 
\begin{array}{c}
\left( c_{1},0.2,-0.5\right) ,\left( c_{2},0.3,-0.6\right) , \\ 
\left( c_{3},0.4,-0.3\right) ,\left( c_{4},0.7,-0.2\right) 
\end{array}%
\right\} 
\end{equation*}%
\ 
\begin{equation*}
F\left( e_{2}\right) =\left\{ 
\begin{array}{c}
\left( c_{1},0.1,-0.6\right) ,\left( c_{2},0.3,-0.5\right) , \\ 
\left( c_{3},0.6,-0.1\right) ,\left( c_{4},0.4,-0.4\right) 
\end{array}%
\right\} 
\end{equation*}%
\begin{equation*}
F\left( e_{3}\right) =\left\{ 
\begin{array}{c}
\left( c_{1},0.2,-0.8\right) ,\left( c_{2},0.4,-0.3\right) , \\ 
\left( c_{3},0.5,-0.3\right) ,\left( c_{4},0.7,-0.3\right) 
\end{array}%
\right\} 
\end{equation*}%
Hence $F\left( e_{1}\right) $, $F\left( e_{2}\right) $ and $F\left(
e_{3}\right) $ are the elements of bipolar fuzzy soft set over a universe $U.
$
\end{example}

\section{An Application of Bipolar Fuzzy Soft Sets in Decision Making}

Bipolar fuzzy soft set has several applications to deal with uncertainties
from our different kinds of daily life problems. Here, we discuss such an
application for solving a socialistic decision making problem. We apply the
concept of bipolar fuzzy soft set for modelling of a socialistic decision
making problem and then we give an algorithm for the choice of optimal
object based upon the available sets of information.

Suppose that $U=\left\{ c_{1}\text{, }c_{2}\text{, }c_{3}\text{, }%
c_{4}\right\} $ be the set of four cars under consideration \ say $U$ is an
initial universe and $E=$\{$e_{1}=$costly, $e_{2}=$Beautiful, $e_{3}=$Fuel
efficient, $e_{4}=$Modern technology, $e_{5}=$Luxurious\} be a set of
parameters.

Suppose a man Mr. $X$ is going to buy a car on the basis of his wishing
parameter among the listed above. Our aim is to find out the attractive car
for Mr. $X$.

Suppose the wishing parameters of Mr. $X$ be $A\subset E$ where $A=\left\{
e_{1}\text{, }e_{2}\text{, }e_{5}\right\} $. Consider the bipolar fuzzy soft
set as below.%
\begin{equation*}
F\left( e_{1}\right) =\left\{ 
\begin{array}{c}
\left( c_{1},0.4,-0.5\right) ,\left( c_{2},0.6,-0.3\right) , \\ 
\left( c_{3},0.8,-0.2\right) ,\left( c_{4},0.5,-0.2\right) 
\end{array}%
\right\} 
\end{equation*}%
\begin{equation*}
F\left( e_{2}\right) =\left\{ 
\begin{array}{c}
\left( c_{1},0.5,-0.5\right) ,\left( c_{2},0.3,-0.1\right) , \\ 
\left( c_{3},0.4,-0.4\right) ,\left( c_{4},0.7,-0.3\right) 
\end{array}%
\right\} 
\end{equation*}%
\begin{equation*}
F\left( e_{5}\right) =\left\{ 
\begin{array}{c}
\left( c_{1},0.7,0\right) ,\left( c_{2},0.5,-0.3\right) , \\ 
\left( c_{3},0.6,-0.3\right) ,\left( c_{4},0.4,-0.4\right) 
\end{array}%
\right\} 
\end{equation*}

\begin{definition}
( Comparison table). It is a square table in which number of rows and number
of columns are equal and both are labeled by the object name of the universe
such as $c_{1}$, $c_{2}$, $c_{3}$,...,$c_{n}$ and the entries $d_{ij}$ where 
$d_{ij}=$ the number of parameters for which the value of $d_{i}$ exceeds or
equal to the value of $d_{j}$.
\end{definition}

\textbf{Algorithm}.

\begin{enumerate}
\item Input the set $A\subset E$ of choice of parameters of Mr. X.

\item Consider the bipolar fuzzy soft set in tabular form.

\item Compute the comparison table of positive information function and
negative information function.

\item Compute the positive information score and negative information score.

\item Compute the final score by subtracting positive information score from
negative information score.
\end{enumerate}

Find the maximum score, if it occurs in i-th row, then Mr. X will buy to $%
d_{i}$, $1\leq i\leq 4$.%
\begin{equation*}
\begin{tabular}{|l|l|l|l|}
\hline
$.$ & $e_{1}$ & $e_{2}$ & $e_{5}$ \\ \hline
$c_{1}$ & $0.4$ & $0.5$ & $0.7$ \\ \hline
$c_{2}$ & $0.6$ & $0.3$ & $0.5$ \\ \hline
$c_{3}$ & $0.8$ & $0.4$ & $0.6$ \\ \hline
$c_{4}$ & $0.5$ & $0.7$ & $0.4$ \\ \hline
\end{tabular}%
\end{equation*}%
Table 1. Tabular representation of positive information function.%
\begin{equation*}
\begin{tabular}{|l|l|l|l|l|}
\hline
$.$ & $c_{1}$ & $c_{2}$ & $c_{3}$ & $c_{4}$ \\ \hline
$c_{1}$ & $3$ & $2$ & $2$ & $1$ \\ \hline
$c_{2}$ & $1$ & $3$ & $0$ & $2$ \\ \hline
$c_{3}$ & $1$ & $3$ & $3$ & $2$ \\ \hline
$c_{4}$ & $2$ & $1$ & $1$ & $3$ \\ \hline
\end{tabular}%
\end{equation*}%
Table 2. Comparison table of the above table.%
\begin{equation*}
\begin{tabular}{|l|l|l|l|}
\hline
$.$ & $\text{Row sum(a)}$ & $\text{Column sum(b)}$ & $\text{Membership
score(a-b)}$ \\ \hline
$c_{1}$ & $8$ & $7$ & $1$ \\ \hline
$c_{2}$ & $6$ & $9$ & $-3$ \\ \hline
$c_{3}$ & $9$ & $6$ & $3$ \\ \hline
$c_{4}$ & $7$ & $8$ & $-1$ \\ \hline
\end{tabular}%
\end{equation*}%
Table 3. Membership score table.%
\begin{equation*}
\begin{tabular}{|l|l|l|l|}
\hline
$.$ & $e_{1}$ & $e_{2}$ & $e_{5}$ \\ \hline
$c_{1}$ & $-0.5$ & $-0.5$ & $0$ \\ \hline
$c_{2}$ & $-0.3$ & $-0.1$ & $-0.3$ \\ \hline
$c_{3}$ & $-0.2$ & $-0.4$ & $-0.3$ \\ \hline
$c_{4}$ & $-0.2$ & $-0.3$ & $-0.4$ \\ \hline
\end{tabular}%
\end{equation*}%
Table 4. Tabular representation of negative information function.%
\begin{equation*}
\begin{tabular}{|l|l|l|l|l|}
\hline
$.$ & $c_{1}$ & $c_{2}$ & $c_{3}$ & $c_{4}$ \\ \hline
$c_{1}$ & $3$ & $2$ & $2$ & $2$ \\ \hline
$c_{2}$ & $1$ & $3$ & $2$ & $1$ \\ \hline
$c_{3}$ & $1$ & $2$ & $3$ & $2$ \\ \hline
$c_{4}$ & $1$ & $2$ & $2$ & $3$ \\ \hline
\end{tabular}%
\end{equation*}%
Table 5. Comparison table of the above table.%
\begin{equation*}
\begin{tabular}{|l|l|l|l|}
\hline
$.$ & $\text{Row sum(c)}$ & $\text{Column sum(d)}$ & Non-m$\text{embership
score(c-d)}$ \\ \hline
$c_{1}$ & $9$ & $6$ & $3$ \\ \hline
$c_{2}$ & $7$ & $9$ & $-2$ \\ \hline
$c_{3}$ & $8$ & $9$ & $-1$ \\ \hline
$c_{4}$ & $8$ & $8$ & $0$ \\ \hline
\end{tabular}%
\end{equation*}%
Table 6. Final Score table.%
\begin{equation*}
\begin{tabular}{|l|l|l|l|}
\hline
$.$ & Positive information$\text{ score(m)}$ & $\text{Negative information
score(n)}$ & $\text{Final score(m-n)}$ \\ \hline
$c_{1}$ & $1$ & $3$ & $-2$ \\ \hline
$c_{2}$ & $-3$ & $-2$ & $-1$ \\ \hline
$c_{3}$ & $3$ & $-1$ & $4$ \\ \hline
$c_{4}$ & $-1$ & $0$ & $-1$ \\ \hline
\end{tabular}%
\end{equation*}%
Table\ 7. Final score table.

Clearly the maximum score is $4$ scored by the car $c_{3}$.

\textbf{Decision:} Mr.\ X will buy $c_{3}$. If he does not want to buy $%
c_{3} $ due to certain reason, his second choice will be $c_{2}$ or $c_{4}$,
so the score of $c_{2}$ or $c_{4}$ are same.

\section{Conclusion}

We combine the concept of bipolar fuzzy set and soft set to introduced the
concept of bipolar fuzzy soft sets. We examine some operations on bipolar
fuzzy softs. We study basic operations on bipolar fuzzy soft set. We define
exdended union, intersection of two bipolar fuzzy soft set. We also give an
application of bipolar fuzzy soft set into decision making problem. We give
a general algorithm to solved decision making problems by bipolar fuzzy soft
set. Therefore, this paper gives an idea for the beginning of a new study
for approximations of data with uncertainties. We will focus on the
following problems : bipolar fuzzy soft relations, bipolar fuzzy soft
matrix, bipolar fuzzy soft function and bipolar fuzzy soft graphs, and
applications in artificial intelligence and general systems.

\end{document}